# Compressed sensing of astronomical images: orthogonal wavelets domains

Vasil Kolev*

***Abstract:*** *A simple approach for orthogonal wavelets in compressed sensing (CS) applications is presented. We compare efficient algorithm for different orthogonal wavelet measurement matrices in CS for image processing from scanned photographic plates (SPP). Some important characteristics were obtained for astronomical image processing of SPP. The best orthogonal wavelet choice for measurement matrix construction in CS for image compression of images of SPP is given. The image quality measure for linear and nonlinear image compression method is defined.*

## INTRODUCTION

Compressed sensing (CS) is a new technique in signal processing, coding and information theory. The goal is to achieve dimensionality reduction to a low-dimensional representation of a signal from which a faithful approximation can be recovered. The CS theory has been successfully applied to various signal processing areas including image processing in astronomy.

It is important that the property of wavelets indicates singularity in images by a large wavelet coefficient, as well as a smooth region by a small wavelet coefficient. This means annulment, very often this is threshold, of small wavelet coefficients and separating of detected celestial objects from the image background. This advantage is included in the CS method by introducing a measurement matrix. The algorithm in [15] used the wavelet transform into the contourlet domain to reduce the size of random measurement matrices as well as improved recovered performance.

One of the original breakthroughs in CS [4, 6, 10, and 12] was to show that methods can be used to efficiently reconstructing the data signal with high accuracy. Since then many alternative methods have been proposed as a faster or superior (in terms of reconstruction rate) alternative to these linear programming algorithms [2]. Also a suite of thresholding based algorithms, either hard thresholding [3] or soft thresholding [9] has been proposed. They represent the CS encoding matrix as a graphical model with fast computation, which is obtained by reducing the size of the graphical model with sparse encoding matrices [8]. In [13] an application on CS for astronomical images is given. The CS algorithm for some image processing of astronomical images is provided in [14]. According to the CS theory, the measurement matrices need to be incoherent [4]. Such matrices are Gaussian random matrix [9], matrices generated from the no binary orthogonal transform matrices [7, 11], matrices generated from the binary orthogonal transform matrices [5], and sparse matrices, which are usually generated from block diagonal matrices [12].

Based CS method image compression of SPP including measurement matrices from different orthogonal wavelets is considered. We do image quality analysis of CS algorithm for image processing from scanned photographic plates (SPP).

* This work has been supported by the research project D0-02-275 of the Institutes of Astronomy with NAO, IMI, and IICT, BAS, Bulgaria.

## INTRODUCTION IN COMPRESSED SENSING

If signal $x$ in $R^J$ ( $J$ -real vector space) is sufficiently sparse we perform reconstruction from $y = \mathbf{A}x$ in $R^I$ for integer $(I << J)$ ( $I$ mush smaller than $J$ ) with sparse matrix $\mathbf{A}$ only $\mathrm{O}(I \log\left(J/I\right))$ linear measurements. In the general case, a minimization problem with cost function is considered:

$$\Im(x) = \arg\min_{x \geq 0} \left\| y - Ax \right\|_2^2 + \lambda J(x)$$

where $\lambda$ is regularization parameter, $\left\| x \right\|_2^2$ - matrix norm, and $J(x)$ - a cost function for sparseness. The function can be easily modified to correspond to the CS definition. Therefore, if there is a sparse signal it is possible to take fewer measurements of the signal and reconstruct the original signal, using optimization techniques. The CS algorithm of image $\mathbf{S}$, obtained from compressed column signals $\mathbf{Y} = [\mathrm{y}_1, \mathrm{y}_2, .., \mathrm{y}_T] \in R^{IxT}$ is presented:

$$\mathbf{Y} = \mathbf{A} * \mathbf{\Psi} * \mathbf{S} \tag{1}$$

where the matrix $\mathbf{S} = [\mathrm{s}_1, \mathrm{s}_2, \mathrm{s}_3, .., \mathrm{s}_T] \in \mathrm{R}^{\mathrm{JxT}}$ represents $T$ columns from the image with $J$ pixel values, and the projection matrix $\mathbf{A} = [\mathrm{a}_1, \mathrm{a}_2, .., \mathrm{a}_J] \in R^{IxJ}$ . The measurement matrix $\mathbf{\Psi}$ is orthogonal transform matrix obtained by DCT, orthogonal wavelets or curvelets which is able to sparsity each signal $\mathrm{s}_\mathrm{t}$ on its domain. In distributed CS the original image has similar properties which are expressed by the dictionary $\mathbf{\Psi}$, which means that the projections are the same for all $\mathrm{s}_\mathrm{t}$. The projection matrix A can be selected adaptively to the dictionary $\mathbf{\Psi}$ or generated by randomly sampling the columns, with different columns independently identically distributed (i.i.d.) on the unit sphere $S^{I-1}$ in Euclidean $I$ -space.

## ALGORITHM FOR COMPRESSED SENSING IN ORTHOGONAL WAVELET DOMAIN

Let us to express the compressed signal (1) by the sparse matrix $\mathbf{X}$:

$$\mathbf{Y} = \mathbf{A} * \mathbf{X} \tag{2a}$$

where $\mathbf{X}$ is defined by:

$$\mathbf{X} = \mathbf{\Psi} * \mathbf{S} \tag{2b}$$

Hence, the image $\mathbf{S}$ has been reconstructed from the inverse transform $\mathbf{S} = \mathbf{\Psi}^\mathbf{T} * \mathbf{X}$. A matrix-vector product (2b) is computed with columns of $\mathbf{\Psi}$ which are orthonormal wavelet bases. Then the multiplications $\mathbf{\Psi v}$ and/or $\mathbf{\Psi}^\mathbf{T}\mathbf{v}$ can be performed by the fast wavelet transforms. Since it is impossible to process directly the whole matrices $\mathbf{A}$ and $\mathbf{X}$, we cannot recover the images based on their compressed versions with global learning rules. We want to form a set of local learning rules, which allow us to perform the estimation sequentially, row-by-row, that is why we will use the product $\mathbf{A}^\mathbf{T}\mathbf{A}$ . After applying of matrix $\mathbf{A}^\mathbf{T}$ for both sides (2a), signal $\mathbf{Y}$ will be represented as:

$$\mathbf{A}^\mathbf{T} * \mathbf{Y} = \mathbf{A}^\mathbf{T} * \mathbf{A} * \mathbf{X} . \tag{3}$$

Let us denote the modified $\mathbf{Y}$ by $\tilde{\mathbf{Y}} = \mathbf{A}^\mathbf{T} * \mathbf{Y}$ , and matrix multiplier $\mathbf{G} = \mathbf{A}^\mathbf{T}\mathbf{A}$ . Due to (2b) we express (3) such as:

$$\tilde{\mathbf{Y}} = \mathbf{G} * \mathbf{X} = \mathbf{G} * \mathbf{\Psi} * \mathbf{S} \tag{4}$$

where the vector matrices are $\tilde{\mathbf{Y}} = [\tilde{\mathbf{y}}_1, \tilde{\mathbf{y}}_2, .., \tilde{\mathbf{y}}_\mathbf{T}] \in \mathbf{R}^{\mathbf{JxT}}$, $\mathbf{G} = [\mathbf{g}_1, \mathbf{g}_2, .., \mathbf{g}_\mathbf{J}] \in \mathbf{R}^{\mathbf{JxJ}}$ . Therefore the error function from (3) is:

$$\mathbf{E} = \mathbf{A}^\mathbf{T} * \mathbf{Y} - \mathbf{A}^\mathbf{T} * \mathbf{A} * \mathbf{X} = \tilde{\mathbf{Y}} \text{ - } \mathbf{GX} \tag{5}$$

for $J$ pixel values. The object is obtain of the matrix $\mathbf{X}$ from $\tilde{\mathbf{Y}}$ and $\mathbf{G}$ .

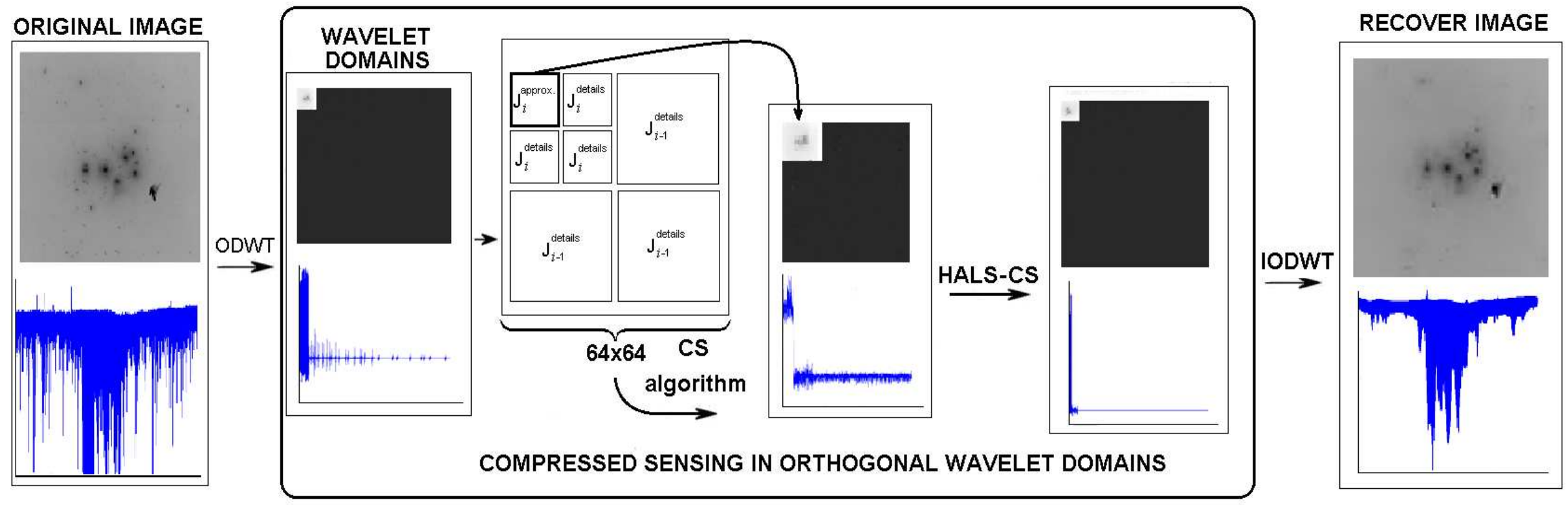

Fig.1 Wavelet domain of CS algorithm for image compression of astronomical image from the SPP **M45-S40-4154p** (512x512 pixels);
(Under each images in the 1D pixels are given)

To find a solution we introduce $j$- divergences between two nonnegative sets of regularized cost functions which minimize the squared Euclidean distances [7]:

$$\mathbf{D}^{\mathbf{j}}(\mathbf{y}^{\mathbf{j}} \,\|\, \mathbf{x}_{\mathbf{j}}) = \frac{\mathbf{1}}{\mathbf{2}}\left\|\mathbf{y}^{\mathbf{j}} - \mathbf{x}_{\mathbf{j}}\right\|_2^2 + \sum_{\mathbf{t}} \boldsymbol{\lambda}_{\mathbf{t}} \mathbf{J}_{\mathbf{x}}(\mathbf{x}_{\mathbf{t}}) \text{ for } \mathbf{j} = \mathbf{1,2,...,J} \qquad (6)$$

where $\mathbf{y}^{\mathbf{j}} = \mathbf{x}_{\mathbf{j}} + \tilde{\mathbf{y}}_{\mathbf{j}} - \mathbf{g}_{\mathbf{j}}\mathbf{X} = \mathbf{x}_{\mathbf{j}} + \mathbf{e}_{\mathbf{j}}$, for $j = 1,2,...,J$, vector $x_t$ are the $t$-column of $\mathbf{X}$, $\boldsymbol{\lambda} = [\boldsymbol{\lambda}_{\mathbf{1}}, \boldsymbol{\lambda}_{\mathbf{2}},..,\boldsymbol{\lambda}_{\mathbf{T}}] \geq \mathbf{0}$ is row residual vector of regularization coefficients, and $\mathbf{J}_{\mathbf{x}}(\mathbf{x}_{\mathbf{t}})$ - a cost function for sparseness or regularization terms with additional constraints.

In Fig.1 the orthogonal wavelet-domain for CS scheme with their image values (in 1D) is shown. Since the random matrix A is generated with uniform spherical ensemble, the $n$-vectors columns are independently from a uniform distribution in Euclidean space, as well as uniformly distributed on the sphere $S^{n-1}$ [10].

I order to compare the image quality with pixel numbers we define an **image reduction level** (IRL) for $i$-level image decomposition as:

$$\mathbf{IRL}_{\mathbf{i}}^{\mathbf{Linear}} = \frac{\textit{pixels of the original image}}{\textit{pixels of i-level image decomposition}} \qquad (7)$$

Two **image compression methods -** linear and nonlinear – exist. The linear one is when $\mathbf{IRL}_i$ **is** equal of order by two. Example, if the image is $\mathbf{256 \times 256}$ after two level decompositions the compressed image obtained 64x64 pixels. Then from (7) calculate that $\mathbf{IRL}_{\mathbf{2}} = \frac{256 \times 256}{64 \times 64} = 16$. Therefore, if image size has been power by two that the **linear image compression method** there is $\mathbf{IRL}$:

$$\mathbf{IRL}_i = 4^i \qquad (8)$$

The CS algorithm is a method using mainly **nonlinear image compression**:

$$\mathbf{IRL}_{\mathbf{i}}^{\mathbf{CS}} = \frac{\textit{pixels of the original image}}{\mathbf{I}_{\mathbf{cs}}} \qquad (9)$$

where $\mathbf{I}_{\mathbf{cs}}$ is the obtained compressed sensing image. From (8) for $i = 4,5$ level decompositions we conclude that $\mathbf{IRL}$s are between $\mathbf{IRL}_4 = 256$ and $\mathbf{IRL}_5 = 1024$. The exact values will be estimated following (10). The factor $\mathbf{IRL}$ characterization reduction level of pixel numbers in CS method with measurement matrix consists from orthogonal wavelets. It can be visual quality tool for image processing, as well as for all CS methods.

==========================================================================

**Algorithm 1: CS algorithm in the orthogonal wavelet domains**

==========================================================================

1. Orthogonal discrete wavelet transformation (ODWT) of image **J** (2b), (Fig.1).
   a) If image size is ($512^2$) ODWT up to level 4.
   b) If image size is ($1024^2$) ODWT up to level 5.
2. Hold approximation coefficients $\mathrm{J}^{\text{aproximate}} = 16 \times 16 = 256$ pixels.
3. Construct a $\mathbf{X}$ - vector of detail coefficients:
   - For level 4 – tree the 16x16 submatrices form $\mathbf{X_4}$ vector with length 768 pixels;
   - For level 5 – tree the 32x32 submatrices form $\mathbf{X_5}$ vector with length 3072 pixels;
4. Compress the above vector with $\mathrm{RR}_4 = 0.75$
   a) Construct randomly sampling the columns forms two A matrices with follows size:
   - For level 4 – the matrix $A_4$ with $\mathrm{RR}_4 * 768 \times 768$ pixels;
   - For level 5 – the matrix $A_5$ with $\mathrm{RR}_5 * 3072 \times 3072$ pixels;

   b) Multiply a vector of detail coefficients with matrix **A:**
   - For level 4 – $\mathbf{Y_4 = A_4 * X_4}$
   - For level 5 – $\mathbf{Y_5 = A_5 * X_5}$
5. Perform steps 3 and 4 for detail coefficients at scale 5 with $\mathrm{RR}_5 = 0.75$.
6. Store the total $\mathbf{J_{cs}}$ compressed sensing image, with length $\mathbf{I_{cs}}$ vector such that (10):

$$\mathbf{I_{cs}} = \mathrm{J}^{\text{approximate}} + \mathrm{RR}_4 \cdot length(\mathbf{J}_4^{\text{details}}) + \mathrm{RR}_5 \cdot length(\mathbf{J}_5^{\text{details}})$$

7. From vectors $\mathbf{Y_4}$ and $\mathbf{Y_5}$ by the learning rule $X(j,:) = sign(X(j,:))\sqrt{X(j,:)^2 - \lambda^2}$ **(**Abramovich shrinkage function) full size submatrices $\mathbf{J}_4^{\text{details}}$ and $\mathbf{J}_5^{\text{details}}$ are constructed.
8. Based inverse ODWT from $\mathbf{J}_4^{\text{details}}$, $\mathbf{J}_5^{\text{details}}$ and $\mathrm{J}^{\text{aproximate}}$ reconstructed image is obtained.

==========================================================================

The Image reconstructed by using Hierarchical Alternating Least Squares (HALS) algorithm [8] based learning rule of Abramovich function [1, 8]:

$$\mathrm{P}_\lambda(X) = sign(X)\sqrt{X^2 - \lambda^2}, |X| \geq \lambda$$

for 10 iterations is made. The threshold values $\lambda$ for each source x is calculated with adaptive linear decreasing strategy:

$$\lambda^{(k)} = \alpha^{(k)} median(\|X(j,:) - meadian(X(j,:))\|$$

$$\alpha^{(k+1)} = \alpha^{(k)} - \Delta_\alpha$$

where $\lambda^0 = \lambda_{\max}$ is first threshold, $k$ is $k$-iteration index, $\Delta_\alpha$ is a decrease step to enforce $\lambda^{(k)}$ towards $\lambda_{\min}$, which is often 0, and $median(\|X(j,:) - meadian(X(j,:))\|)$, which is median absolute deviation function. Since a small number $k$ - iterations given a poor performance while a large number - computationally expensive the threshold should be determinate with decreasing λ. This is key factor to obtain a good performance. First of all we set the iteration index $k = k_{\max}$. The next to form a rough shape for $x$ from extract significant coefficients we set the threshold $\lambda = \lambda_{\max}$. Since values $x$ are sparse, then the coefficients distribution will decrease fast at large intensities, and more slowly for small ones. Therefore, in first iterations threshold $\lambda$ should also decrease fast. Then we can vary threshold $\lambda$ slowly tend to zero. In the end we obtained successive iterations and close to perfect reconstruction.

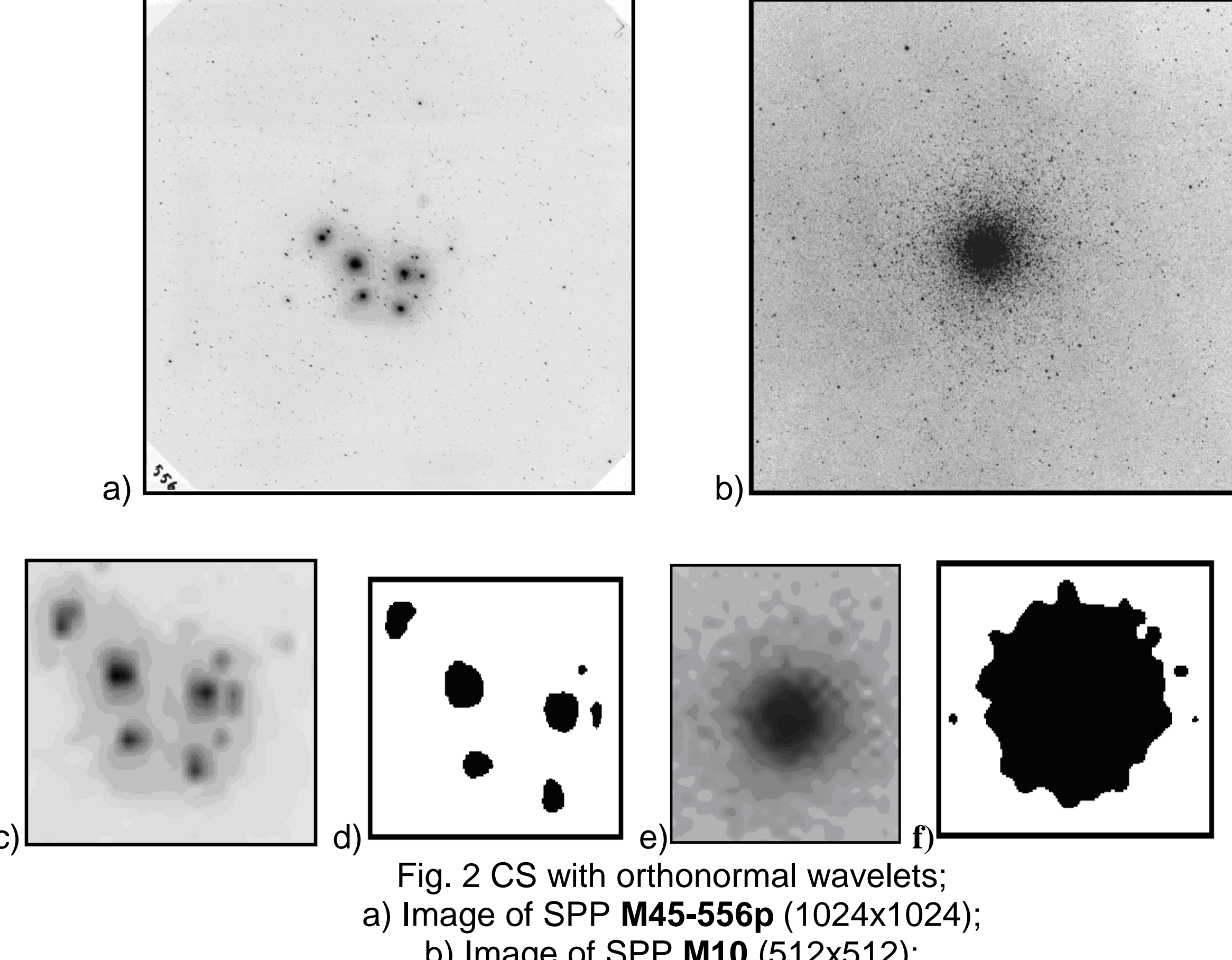

Fig. 2 CS with orthonormal wavelets;
a) Image of SPP **M45-556p** (1024x1024);
b) Image of SPP **M10** (512x512);
A zoomed central area of CS images with 3136 coefficients:
c) Daubechies–4 with $\varepsilon = 0.0098$, $PSNR = 31.13, dB$ and $\mathbf{IRL}_4^{\mathbf{CS}} \approx 334$
d) Celestial object detections of c)
e) Coiflet-30 with $\varepsilon = 0.0107$, $PSNR = 23.71\, dB$, and $\mathbf{IRL}_4^{\mathbf{CS}} \approx 84$
f) Celestial object detections of e)

For better comparison we give two measures to image quality. The first one is the peak-signal-to-noise-ratio $\mathrm{PSNR} = 10\log_{10}\left(255^2/\mathrm{MSE}\right), \mathrm{dB}$ with mean square error $\mathrm{MSE} = \left(1/(\mathrm{m}\times\mathrm{n})\right)\left(\sum_{\mathrm{i}=1}^{\mathrm{M}}\sum_{\mathrm{j}=1}^{\mathrm{N}}(\mathrm{X}(\mathrm{i},\mathrm{j}) - \mathrm{Y}(\mathrm{i},\mathrm{j}))^2\right)$, where pixel values $\mathrm{X}(\mathrm{i},\mathrm{j})$ and $\mathrm{Y}(\mathrm{i},\mathrm{j})$ in the position $(i.j)$, and $\mathrm{M},\mathrm{N}$ are the width and the height of image. The second measure of image quality is the recover error $\varepsilon = \left\|\mathrm{X} - \tilde{\mathrm{X}}\right\|_2 / \left\|\mathrm{X}\right\|_2$, where $\tilde{\mathrm{X}}$ is the image.

## APPLICATIONS OF COMPRESSED SENSING FOR ASTRONOMICAL IMAGES OF SPP

First, let to consider $\mathbf{IRL}$ for images with 512x512 pixels. The length of obtained compressed image is:

$$\mathbf{I_{cs}} = \mathrm{J}^{\mathrm{approximate}} + \mathrm{RR}_4 \cdot \mathrm{J}_4^{\mathrm{details}} + \mathrm{RR}_5 \cdot \mathrm{J}_5^{\mathrm{details}} = (2^4)^2 + \mathrm{RR}_4 \cdot 3(2^4)^2 + \mathrm{RR}_5 \cdot 3(2^5)^2 \qquad (10)$$

where $\mathrm{J}^{\mathrm{approximate}}$ are number pixels of approximation image, $\mathrm{J}_4^{\mathrm{details}}$ are number pixels of the detail images of 4-level, $\mathrm{J}_5^{\mathrm{details}}$ are number pixels of the detail images of 5-level, and $\mathrm{RR}_i = [0 \div 1]$ is $i-$level reduction rate.

**Table 1**

| | **images from Internet** | | **astronomical images from scanned photographical plates** | | | |
|---|---|---|---|---|---|---|
| **Orthogonal Wavelets**<br>(PSNR in dB Wavelet name - filter taps ) | **Lena**<br>(512x512) | **Mondrain**<br>(512x512) | **M10**<br>(512x512) | **ADH5269**<br>(512x512) | **ROZ200001655a**<br>(1024x1024) | **M45-556p**<br>(1024x1024) |
| **Beylkin–18** | **psnr = 22.27**<br>$\mathcal{E}=$**0.0396** | **psnr = 20.17**<br>$\mathcal{E}=$**0.0922** | **psnr = 23.71**<br>$\mathcal{E}=$**0.0107** | **psnr =19.87**<br>$\mathcal{E}=$**0.0216** | **psnr = 24.88**<br>$\mathcal{E}=$**0.0145** | **psnr = 30.64**<br>$\mathcal{E}=$**0.0094** |
| **Coiflet-6** | **psnr = 22.63**<br>$\mathcal{E}=$**0.0368** | **psnr = 20.37**<br>$\mathcal{E}=$**0.0930** | **psnr = 23.71**<br>$\mathcal{E}=$**0.0108** | **psnr = 19.94**<br>$\mathcal{E}=$**0.0212** | **psnr = 25.01**<br>$\mathcal{E}=$**0.0110** | **psnr = 30.91**<br>$\mathcal{E}=$**0.0106** |
| **Coiflet -30** | **psnr = 22.92**<br>$\mathcal{E}=$**0.0364** | **psnr = 20.24**<br>$\mathcal{E}=$**0.0914** | **psnr = 23.71**<br>$\mathcal{E}=$**0.0107** | **psnr = 20.00**<br>$\mathcal{E}=$**0.0215** | **psnr = 24.96**<br>$\mathcal{E}=$**0.0132** | **psnr = 30.92**<br>$\mathcal{E}=$**0.0103** |
| **Daubechies-4** | **psnr = 22.14**<br>$\mathcal{E}=$**0.0389** | **psnr = 20.21**<br>$\mathcal{E}=$**0.0927** | **psnr = 23.70**<br>$\mathcal{E}=$**0.0107** | **psnr = 20.00**<br>$\mathcal{E}=$**0.0214** | **psnr = 24.97**<br>$\mathcal{E}=$**0.0119** | **psnr = 30.93**<br>$\mathcal{E}=$**0.0097** |
| **Daubechies-16** | **psnr = 22.39**<br>$\mathcal{E}=$**0.0386** | **psnr = 20.31**<br>$\mathcal{E}=$**0.0940** | **psnr = 23.71**<br>$\mathcal{E}=$**0.0108** | **psnr = 19.89**<br>$\mathcal{E}=$**0.0211** | **psnr = 24.90**<br>$\mathcal{E}=$**0.0153** | **psnr = 30.68**<br>$\mathcal{E}=$**0.0094** |
| **Symmlet-8** | **psnr =22.85**<br>$\mathcal{E}=$**0.0364** | **psnr = 20.61**<br>$\mathcal{E}=$**0.0878** | **psnr = 23.71**<br>$\mathcal{E}=$**0.0108** | **psnr = 20.03**<br>$\mathcal{E}=$**0.0210** | **psnr = 25.02**<br>$\mathcal{E}=$**0.0117** | **psnr = *31.13***<br>$\mathcal{E}=$**0.0098** |
| **Vaidyanathan-24** | **psnr = 22.30**<br>$\mathcal{E}=$**0.0379** | **psnr = 20.17**<br>$\mathcal{E}=$**0.0908** | **psnr = 23.70**<br>$\mathcal{E}=$**0.0107** | **psnr = 19.85**<br>$\mathcal{E}=$**0.0216** | **psnr = 24.86**<br>$\mathcal{E}=$**0.0150** | **psnr = 30.56**<br>$\mathcal{E}=$***0.0093*** |

Since $\mathrm{RR}_i = 0.75$ the number of pixels of the recover image (10) are:

$$\mathbf{I}_{\mathbf{cs}} = 256 + \mathrm{RR}_4 \cdot 768 + \mathrm{RR}_5 \cdot 3072 = 256 + \frac{3}{4} \cdot 768 + \frac{3}{4} \cdot 3072 = 3136 \qquad (11)$$

Therefore, that the $\mathbf{IRL}$ for image with 512x512 pixels is $\mathbf{IRL}_4^{\mathbf{CS}} = (512 \times 512)/3126 \approx 84$, for 1024x1024 pixels -$\mathbf{IRL}_4^{\mathbf{CS}} \approx 334$, for 2048x2048 pixels - $\mathbf{IRL}_4^{\mathbf{CS}} \approx 1338$, and 4096x4096 pixels are $\mathbf{IRL}_4^{\mathbf{CS}} \approx 5350$ so on.

The all used orthogonal wavelet filters are included in dictionary $\mathbf{\Psi}$ described in Table1. The experiments were performed with image of SPP (Fig. 1) with CS algorithm [8] presentation in details in algorithm 1. The present is applied for images with 512x512 and 1024x1924 pixels. The obtained results are displayed in Table1. Since are obtained the different recover errors $\varepsilon$, but with equal PSNR, we give both recover errors and PSNR. Example, Lena image has recover error $\varepsilon = 0.0364$ with Symmlet-8 and Coiflet-30, but $PSNR$ is a little bit different. The same result is obtained at Beylkin and Vaidyanathan wavelets for image from SPP **ADH5269**, but $PSNR$ of Beylkin-18 wavelet is better.

The biggest $PSNR$ has Symmlet wavelet with eight coefficients, i.e., $PSNR = 31.13, dB$ with error $\varepsilon = 0.0098$, also for extented central celestial objects, **M45-556p**. Interest is the case for the image of SPP **M10**, where for all orthogonal wavelet families the nearly equal image quality: $PSNR \approx 23.71\,\mathrm{dB}$ with $\varepsilon \approx 0.0108$ are obtained. The same result has been obtained for the image of SPP **M45-556p**. Therefore, images with celestial objects mainly in the image center (Fig.2c and e) are obtained with $PSNR > (3 \div 8)\,\mathrm{dB}$ with error more that 3 times compared to the other images.

Due to a minimal filter support and results (see Table1), best choice is Symmlet8 orthogonal wavelet for CS in images of SPP. The symmetric and smoothness of Symmlet8, Coiflet30 preserve energy and give better $PSNR$ with small error $\varepsilon$. They are useful for images of SPP with celestial objects in the central area (as shown in Fig. 2c-d). The image of SPP **ADH5269** consisting from scattered celestial objects obtains the smallest PSNR with error 2-4 times more than the other images.

Let to consider the image from SPP **M45-556p** (Fig.2a). This is the Pleiades star cluster or the Seven Sisters which contains hundreds of stars. This is one of the brightest star clusters visible in the northern hemisphere and consists of many bright, hot, young stars. The all stars formed at the same time around 100 million years ago within a large cloud of interstellar dust and gas. By using nonlinear shrinkage function we can celestial objects is detected (see Fig. 2c and d). The image fro SPP **M10** (Fig.2b) is Globular cluster lies in the constellation of Ophiuchus. This is a very bright cluster with a central region that appears slightly pear-shaped. It is about 70 light-years in diameter (Fig.2f) and lies about 16,000 light-years from Earth. As well as image from SPP **M45-556p** also by applied of nonlinear shrinkage function we the celestial object is detected (see Fig. 2e and f). This image processing are useful for stars detection, the stellar diameters calculated and catalogs construct at astronomical images.

## CONCLUSIONS AND FUTURE WORK

We present method which leads to simple CS algorithm in orthogonal wavelet domain. It gives small errors and can be used for image compression of SPP. Obtained are the important characteristics for processing with CS method with orthogonal wavelets. The analysis shows that the best choice for image of SPP is wavelets - Symmlet, Daubechies, and Coiflet. The present CS algorithm is especially effective for noise remove. By used nonlinear shrinkage function we can made stars detection. Since CS method is dependent on the image types, processing of images with scatter celestial objects is none appropriate. The image reduction levels for linear and nonlinear image compression are introduces. This is useful for image quality analysis of the CS method.

## ACKNOWLEDGEMENTS

I wish to acknowledge Prof. O. Kounchev from Institute of Mathematics and Informatics, Bulgarian Academy of Sciences for the useful discussions and helpful comments. I am grateful to Prof. K. Tsvetkova and Prof. M. Tsvetkov, from Institute of Astronomy and National Astronomical Observatory, Bulgarian Academy of Sciences, for providing me images of scanned photographic plates from Sofia Sky Archive Data Center, Bulgarian Academy of Sciences. I would like to acknowledge the anonymous referees for making very useful suggestions.

**ABOUT THE AUTHOR**

Vasil Kolev, Institute of Information and Communications Technologies, Bulgarian Academy of Sciences, E-mail: kolev_acad@abv.bg.